# Image2PCI – A Multitask Learning Framework for Estimating Pavement Condition Indices Directly from Images

**Neema Jakisa Owor**
Master's Student
University of Missouri-Columbia
Civil and Environmental Engineering Department
Email: nodyv@umsystem.edu

**Hang Du**
PhD student
University of Missouri-Columbia
Civil and Environmental Engineering Department
Email: dh63r@umsystem.edu

**Abdulateef Daud**
PhD student
University of Missouri-Columbia
Civil and Environmental Engineering Department
Email: aadcvg@umsystem.edu

**Armstrong Aboah**
Assistant Research Professor
University of Arizona
Civil and Architectural Engineering and Mechanics Department
Email: aaboah@arizona.edu

**Yaw Adu-Gyamfi**
Associate Professor
University of Missouri-Columbia
Civil and Environmental Engineering Department
Email: adugyamfi@missouri.edu

Word Count: 6601 words + 1 table (250 words per table) = 6851 words

Submitted for consideration for presentation at the 103rd Annual Meeting of the Transportation Research Board, January 2024

*Submitted [August 1, 2023]*

*Owor, Du, Daud, Aboah and Adugyamfi***ABSTRACT**
The Pavement Condition Index (PCI) is a widely used metric for evaluating pavement performance based on the type, extent and severity of distresses detected on a pavement surface. In recent times, significant progress has been made in utilizing deep-learning approaches to automate PCI estimation process. However, the current approaches rely on at least two separate models to estimate PCI values—one model dedicated to determining the type and extent and another for estimating their severity. This approach presents several challenges, including complexities, high computational resource demands, and maintenance burdens that necessitate careful consideration and resolution. To overcome these challenges, the current study develops a unified multi-tasking model that predicts the PCI directly from a top-down pavement image. The proposed architecture is a multi-task model composed of one encoder for feature extraction and four decoders to handle specific tasks: two detection heads, one segmentation head and one PCI estimation head. By multitasking, we are able to extract features from the detection and segmentation heads for automatically estimating the PCI directly from the images. The model performs very well on our benchmarked and open pavement distress dataset that is annotated for multitask learning (the first of its kind). To our best knowledge, this is the first work that can estimate PCI directly from an image at real time speeds while maintaining excellent accuracy on all related tasks for crack detection and segmentation.

**INTRODUCTION**

Road infrastructure plays a vital role in the development of cities and countries, including determining the comfort and welfare of its users[1]. However, over time, the road surface degrades, leading to adverse effects on the comfort and safety of the road users. Consequently, pavement condition monitoring and evaluation systems have therefore been developed to assess roads and prioritize them for routine maintenance and rehabilitation activities [2]. One such system is the Pavement Condition Index (PCI) which is utilized by transportation agencies to rate and rank pavements for determining their maintenance and rehabilitation needs. Unlike other rating system that may suffice with wind-shield surveys, the PCI requires specific measurements of distress type, extent and severity, making the assessment process tedious, time-consuming, unsafe and cost-prohibitive for a large network of roads. Typically, PCI assessments are recommended only for representative samples, covering 10% of each road segment. Therefore, conducting comprehensive, network level PCI surveys is only feasible through automated means.

The era of artificial intelligence powered by recent advances in deep learning is fueling an increasing demand for fast, low-cost, but accurate network-level pavement condition monitoring. As a result, numerous studies have been conducted into the development of deep-learning distress analysis algorithms to support automated pavement condition monitoring. The general approach to automated distress analysis has however been to use two separate models instead of one. Sarmiento [3] for example used two independent models to detect and segment the distress. Yolov4 was used to detect the distress and used Deep Labv3 to segment the pavement distress. Deeplabv3 was used because it uses Atrous Spatial pooling to solve the challenge of fuzzier boundaries. [4] first passes the images through a crack type classification using squeeze net, then U-Net was used to segment the images, Mobilenet-SSD was then used for object detection. In the post processing, severity of the images was determined. [5] used Mask RCNN to detect the cracks and in the second step, image processing is employed to determine the severity of the crack.

These two-step processes achieve the ultimate goal of PCI estimation as seen in Hamed et al. [6] who proposed a framework that estimates pavement condition index based on two different models: YOLO-based model which outputs crack type-extent and a UNET model for extracting the severity of the cracks. While this yielded accurate PCI estimates, there are several limitations to this approach: first, it is computationally expensive because it requires two separate models to be developed and non-practical for deployment on edge devices for network level pavement condition assessment. Secondly, the two tasks (severity and type-extent) are closely related as such the features extracted from the segmentation task could have been used to improve the accuracy of the YOLO-based model and vice-versa. By training these models separately, we lose the opportunity to leverage information that could be shared to generalize better and perform at higher levels of accuracy for multiple, related tasks. We therefore propose a multitasking model that performs all these closely related tasks in a single step.

The goal of this study is to develop a single multitask model that can estimate PCI directly from an image and introduce a new benchmark of annotated datasets with polygon-based annotations that combine both the type and severity of distresses for training and benchmarking unified Pavement Condition Index (PCI) estimation models.

Multitasking (MTL) is a machine learning approach in which a single model is trained to execute multiple related tasks at the same time. Its benefits include improved generalization due to this shared knowledge, shorter training time and robustness due to underlying structures of the tasks which been realized in[7], [8]. MTL is well-suited for PCI estimation because it requires inputs from closely related tasks: type-extent and severity. The proposed multitasking model (Image2PCI) has one encoder for feature extraction and four decoder heads for specific tasks: two detection heads, one segmentation head and one PCI estimation head. One detection head is to detect the linear distresses and the other head is to detect the pattern distresses. The outputs from the two detection heads and the segmentation head are concatenated to form the PCI head. The detection heads were separated into linear and pattern distresses because the linear cracks (longitudinal and transverse cracks) both carry the same deduct value when estimating the PCI. On the other hand, the pattern cracks (alligator, block and patch) show advanced stages of the linear cracks and generally have higher deduct values when estimating the PCI. In the detection head, the model learns the type of distress and implicitly learns the extent --the density of the pixels within the bounding box. Therefore, from the detection head, the model implicitly learns the type and extent of the distress, two of the components used for PCI estimation. The segmentation is pixel-level classification; therefore, the model is implicitly learning the severity of the crack. Severity of linear cracks is determined by the width of the crack and if the crack is segmented, implicitly the model learns the severity of the crack. For the pattern cracks, the model can implicitly learn the severity of the distress based on the area they cover, where larger areas typically indicate more severe pattern distresses. The features learned from the detection head (type and extent) and the segmentation head(severity), are concatenated to form the PCI head which estimates the PCI. This multi-tasking approach enables our model to accurately estimate the PCI directly from the image.

This paper contributes to the body of research in the following ways:
1. The study introduces a novel multi-task model architecture that efficiently handles various aspects of the distress analysis including crack detection, segmentation and PCI estimation. Our model implicitly learns severity, extent, and crack type simultaneously, reducing computational costs significantly by leveraging shared features from the detection and segmentation heads. Its lightweight design enables deployment on edge devices for real-time pavement condition monitoring, making it highly efficient and accessible on a single node. Remarkably, this model is the first of its kind to directly estimate PCI from images.
2. Open-sourced a unique dataset for training and benchmarking unified PCI estimation models. This dataset contains polygon-based annotations that have combined both the type and severity of the distress. (https://github.com/neemajakisa/Pavement-Distress-Dataset/)

**LITERATURE REVIEW**
Traditional Machine Learning methods for pavement distress classification have been used for over 30 years, and with the discovery of deep learning in 2016, we have seen a significant increase in automatic pavement classification research using deep learning [9]. Some of the traditional Machine learning techniques employed include Support Vector Machine [10]–[14]. Random Forests [15]–[17] and Gradient Boosting [18]–[20] and shallow Artificial Neural Networks [21]–[24]. The drawback with these traditional machine learning methods is that they heavily rely on manually handcrafted features which are time-consuming and subjective. In Contrast, deep learning methods automatically learn features from raw data and hence reduce the reliance of manual feature engineering. Most of the recent research on pavement distress classification, detection or segmentation heavily employ deep learning methods.

**Classification of Pavement Distress**
Classification, the process of categorizing the cracks in an image, can be done by classifying the entire image or by categorizing patches within the image. Gopalakrishnan [14] used truncated pretrained Deep Convolutional Neutral Networks (DCNNs) and Single layer neural network for binary classification of pavement cracks of the entire image. [25], [26] used ensemble models and Vision Transformer respectively for image classification. In Contrast, [27], [28] classified patches in an image. These papers only classify the pavement distresses, however, to conduct a comprehensive assessment of pavement condition, it is important to determine the crack type, severity, and extent. Classification of pavement cracks only provides us with information on the type of crack; patch-classification can provide information about the location of the crack, but this is still insufficient to conduct a comprehensive pavement analysis. This paper seeks to address this gap through a wholistic model that implicitly learns type, extent and severity and estimates the PCI.

**Detection of Pavement Distress**
Detection refers to identifying and locating pavement distress in an image and outputting a bounding box around the distress. Similarly, to [14] the concept of transfer learning on Yolov4 Yolov3, and Faster RNN was used by [29] to detect the six distinct pavement distresses on images captured by Unmanned Aerial

Vehicles (UAVs). New techniques like attention mechanism and multi features fusion have been implored by [30] and [31] on Yolov5 to boost the existing model and it improved the precision of the detection by 5.3% and 4.3% respectively. Similarly, to patch-classification, detection can provide localization of the pavement distress, however, is not sufficient to obtain the severity and the extent of the crack to conduct a comprehensive pavement analysis.

**Segmentation of Pavement Distress**
Segmentation is the process of dividing a digital image into different segments. This was also defined by Koutsopoulos et al [32] as the process of extracting important objects (pavement distresses) from the background. The authors of this study investigated algorithms for segmenting pavement images and assessed their effectiveness in distinguishing distress from background. Koutsopoulos and Downey [33] developed algorithms for image enhancement, segmentation, and distress classification based on statistical model that explains the properties of pavement distresses. While image enhancement eliminates the "average" background, segmentation assign values based on their likelihood of being associated with a particular distress type. The approach used in this study first involves identifying primitives (the building blocks of various distresses), followed by classifying images as distressed depending on the outcomes of the first step. Cheng et al [34] developed a novel method for detecting pavement distress based on fuzzy logic, in which pixels associated with a distress are darker than their surroundings. The image brightness function is used to determine how much darker each pixel is in comparison to its surroundings. The fuzzified image is then mapped to the distress domain using pixel values associated with the distresses. The connectivity of the darker pixels is then examined to eliminate pixels that lack connectivity. Finally, distresses are classified using an image projection method. Similar approach was employed by Hassani and Tehrani [35] to develop an automatic pavement distress detection model. Segmentation is a pixel-wise classification of pavement distress and important features like severity and extent can be estimated from it. Nonetheless, the severity and extent alone cannot be used to give a comprehensive pavement analysis and make informed maintenance decisions.

**Severity of Distresses**
Severity of pavement distress has been calculated using a couple of methods but not limited to; image processing [36], [37], 3D point cloud [38], and stereo vision [39]. Wang et al [40] classified pavement cracks into three severity classes; low, high and no crack according to TB 10005-2015 Standard. He employed the canny edge operators on a binary image to calculate the pixel width of each crack. Annotated images were fed into Inception Res-Net2, to determine the severity of the cracks. In [38], RGB images and depth information were combined using an Intel depth camera. YOLOv3 was employed to detect and classify images into 11 different severities. The region of interest for potholes was extracted and projected onto a 3D point cloud. This allowed the researchers to calculate the depth and diameter of the potholes. The severity of the potholes was determined based on the American Society for Testing and Materials (ASTM) D 6433-11 standard. Guan [39] used stereo vision to determine the pavement distress severity. Using three (3) GoPro Hero8 cameras, he reconstructed the pavement distress images to a 3D image, and the volume of the potholes was calculated from the reconstructed depth image. Liu [41] used infrared technology to classify fatigue cracks into four severity classes: no, low, medium, and high. The severity classification was based on the visual interpretation as stated in the Distress Identification Manual for The Long-Term Pavement Performance Program and Standard Practice for Roads and Parking Lots Pavement Condition Index Survey.

These papers give promising results on estimating the results of the severity of the pavement distress however, they only focused on pavement severity. For us to shift to automatic pavement condition rating, the models developed should be able to detect the type of crack, the severity and the extent of the crack simultaneously.

**Multitask Learning**

Recent implementations of deep Machine learning models have begun introducing multi-task learning (MTL) into model architectures, either implicitly or explicitly. Two main MTL approaches are often used: hard and soft parameter sharing. Hard parameter sharing is used for closely related tasks while soft parameter sharing is used for loosely related tasks. Liang et al [7] for example proposed an MTL approach for computer vision. Their approach shares convolutional layers, while learning task-specific, fully connected layers. Although the structure of sharing is pre-defined, they placed matrix priors on the fully connected layer to enable the model to learn relationship between tasks. Results showed improved performance on well-studied computer vision problems, but error-prone for novel tasks. Other studies explored a bottom-up approach: [8] proposed an alternative MTL approach that considers the uncertainty of each task instead of learning the structure of sharing. The relative weight of each task in the loss function is adjusted by deriving a multi-task loss function based on maximizing the Gaussian likelihood with task-dependent uncertainty. Their architecture performed well when used for simultaneous per-pixel depth regression, semantic and instance segmentation.

MTL is well-suited for PCI estimation because it requires inputs from closely related tasks: type-extent and severity. The general approach to automated distress analysis has however been to use two separate models instead of one. [3]–[5] used two-step processes to determine the type and severity of the pavement distresses. These two-step processes achieve the ultimate goal of PCI estimation as seen in Hamed et al. [6] who proposed a framework that estimates pavement condition index based on two different models: YOLO-based model which outputs crack type-extent and a UNET model for extracting the severity of the cracks. Even though this method yielded accurate PCI estimates, it has a number of limitations: First, it is computationally expensive because two separate models must be created, and it is impractical for deployment on edge devices for network-level pavement condition assessment. Second, since the two tasks (severity and type-extent) are closely related, the segmentation task's extracted features could have been used to enhance the accuracy of the YOLO-based model and vice versa. By training these models independently, we forfeit the opportunity to leverage information that could be shared to improve generalization and achieve higher levels of accuracy for multiple tasks that are interrelated. Therefore, we propose a multitasking model (Image2PCI) that performs all these closely related tasks in a single phase, creating a single model that can estimate PCI directly from an image.

**METHODS**

Three key steps were followed to develop and evaluate the MTL approach for estimating PCI directly from an image as shown in **Figure 1**. First, we annotated and benchmark pavement images for type, extent, severity and PCI, using polygon-based annotation techniques. Second, the architecture of a deep MTL is developed taking advantage of multi-head, encoder-decoder architectures to implicitly learn features for prediction of type, extent, severity simultaneously and estimate the PCI of a pavement image. Third, the developed framework is then evaluated. Detailed description of the methods implemented in this study are discussed as follows:

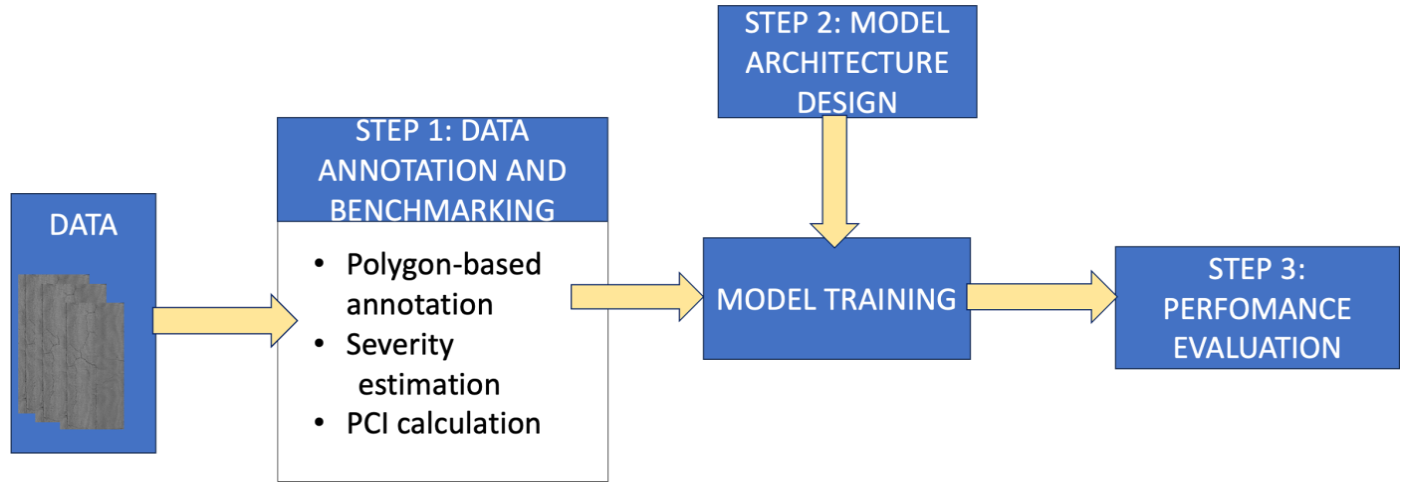

**Figure 1: Methodology**

**Data Annotation and Benchmarking**

*Polygon-based Annotation*
Because this is the first application of MTL for automated distress analysis, none of the existing annotated pavement image databases [42]– [44] were useful. The dataset consists of top-down images of pavements that are obtained from Laser Crack Measurement System. The Computer Vision Annotation Tool (CVAT) was used for the polygon-based annotation of the pavement distresses. Annotators categorized the distresses based on two crucial aspects of PCI: their type and severity. The distress types included alligator, longitudinal, transverse, block, manhole, and patch, whereas the severity levels were classified as low, medium, and high. To ensure the quality and accuracy of the segmentations, civil engineering students who are well-versed with American Standard Testing Materials (ASTM) D6433-11 were involved in the annotation process. 766 images were annotated, and **Table 1** shows the data distribution.

**Table 1: Data distribution**

| | Distress Type distribution | |
|---|---|---|
| | Distress type | Number of Annotations |
| 1. | Alligator | 481 |
| 2. | Block | 265 |
| 3. | Longitudinal | 903 |
| 4. | Patch | 138 |
| 5. | Transverse | 604 |
| | Distress severity distribution | |
| 1. | Low | 178 |
| 2. | Medium | 1466 |
| 3. | High | 509 |

*Severity estimation*

Visual inspection was used to determine the type of pavement distress; however, the use of visual inspection to evaluate the severity of pavement distress is highly subjective. To establish a more standardized approach, the different pavement distresses were classified into low, medium, and high severity levels, in accordance with ASTM D-6433-11 guidelines. The number of pixels in a distress that represent the thresholds of the different severity levels were calculated using the formula in **Equation 1.**

$$pixel = \left(\frac{total\ pixels\ along\ the\ width\ or\ length}{the\ actual\ length\ or\ width\ of\ the\ image\ in\ mm}\right) x\ threshold\ in\ mm. \quad (1)$$

A python computer program was written to calculate the number of white pixels represented by a distress. By clicking on two opposite sides of the distress (shown by the blue and red dots on the **Figure 2**) the program was able to calculate the pixel width in-between the two points. The pixel width was calculated along three separate places along the pavement distress. The average width of these three pixel widths was then taken as the average pixel width of the pavement distress. Following the pixel thresholds calculated, the pavement distress was classified according to its severity level.

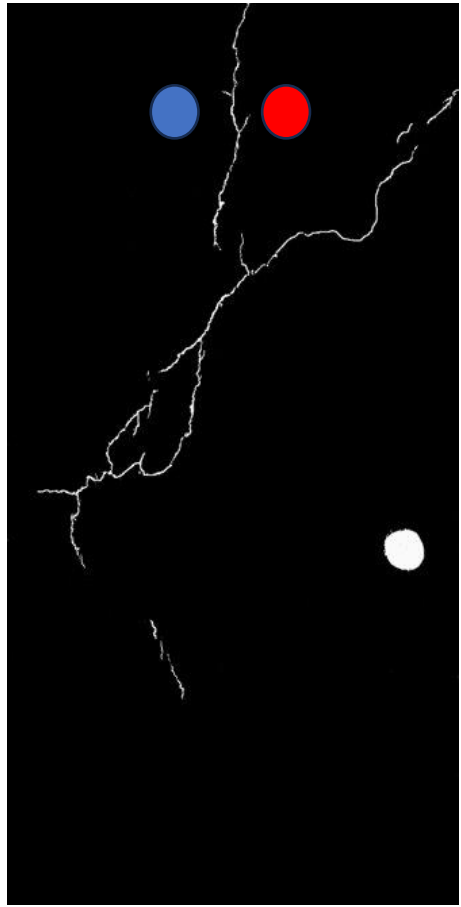

**Figure 2: Image Mask**

*PCI calculation*
The annotations done in CVAT classifications combined the type and severity of the distress as shown in the **Figure 3.** The extent of the pavement distresses was computed from the area of the polygon-annotations. Following the procedure in ASTM-D643-11, a program was written to calculate the PCI of each image.

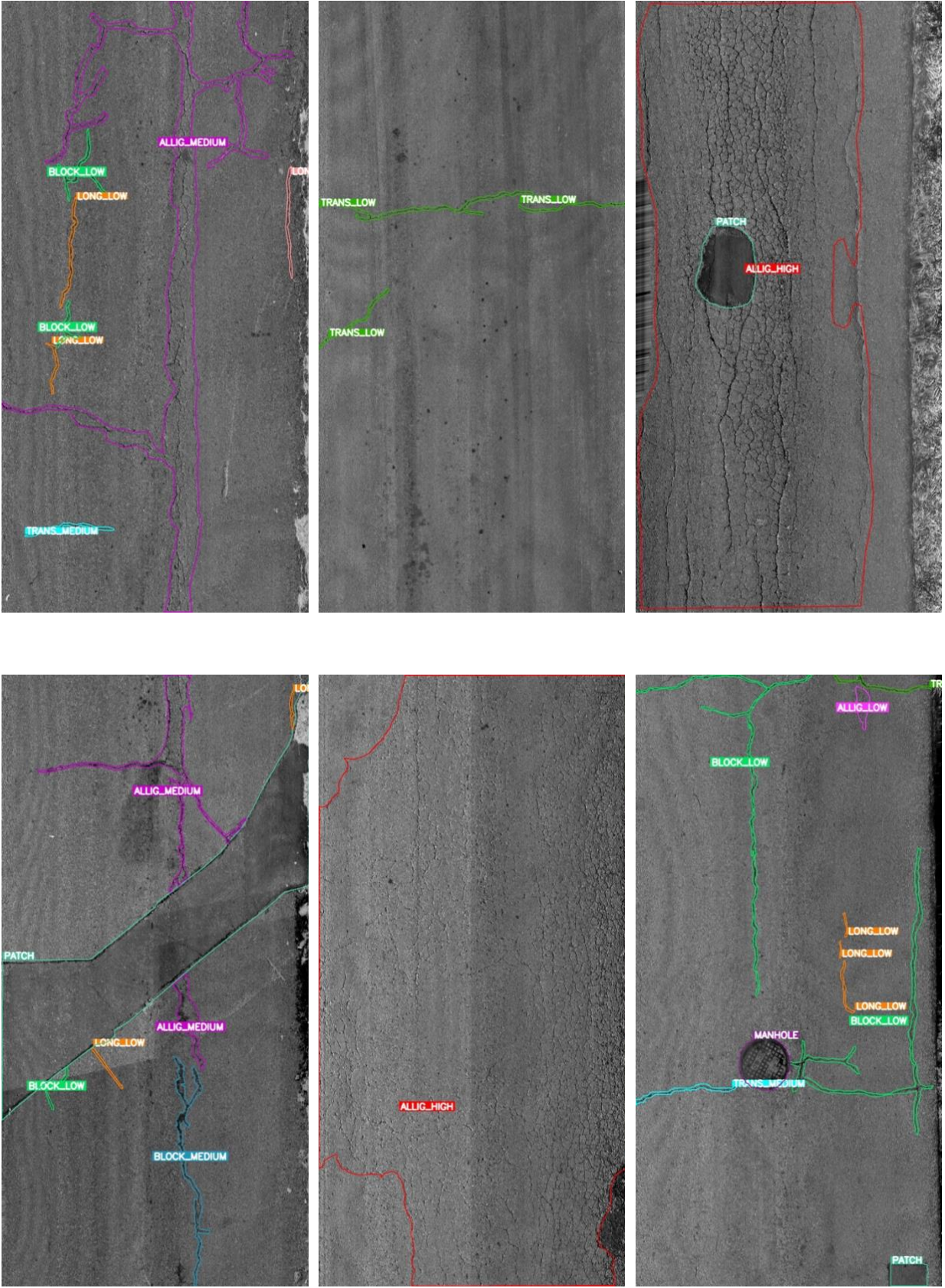

Figure 3: Polygon-annotated dataset

**MTL Architecture**

Image2PCI is a multi-task learning framework that estimates the PCI directly from an image. The model architecture was inspired by [45]. **Figure 4** illustrates our modified network architecture which consists of a single Encoder responsible for extracting features and four Decoder units. These Decoder heads are comprised of two for detection, one for segmentation, and one for PCI estimation. One of the detection heads specializes in learning features associated with linear distresses (longitudinal, transverse cracks), while the other focuses on capturing features associated with pattern distresses (block, alligator, patch). The outputs from the detection heads (linear distress head, pattern distress head) and the segmentation head are then concatenated to form the PCI head.

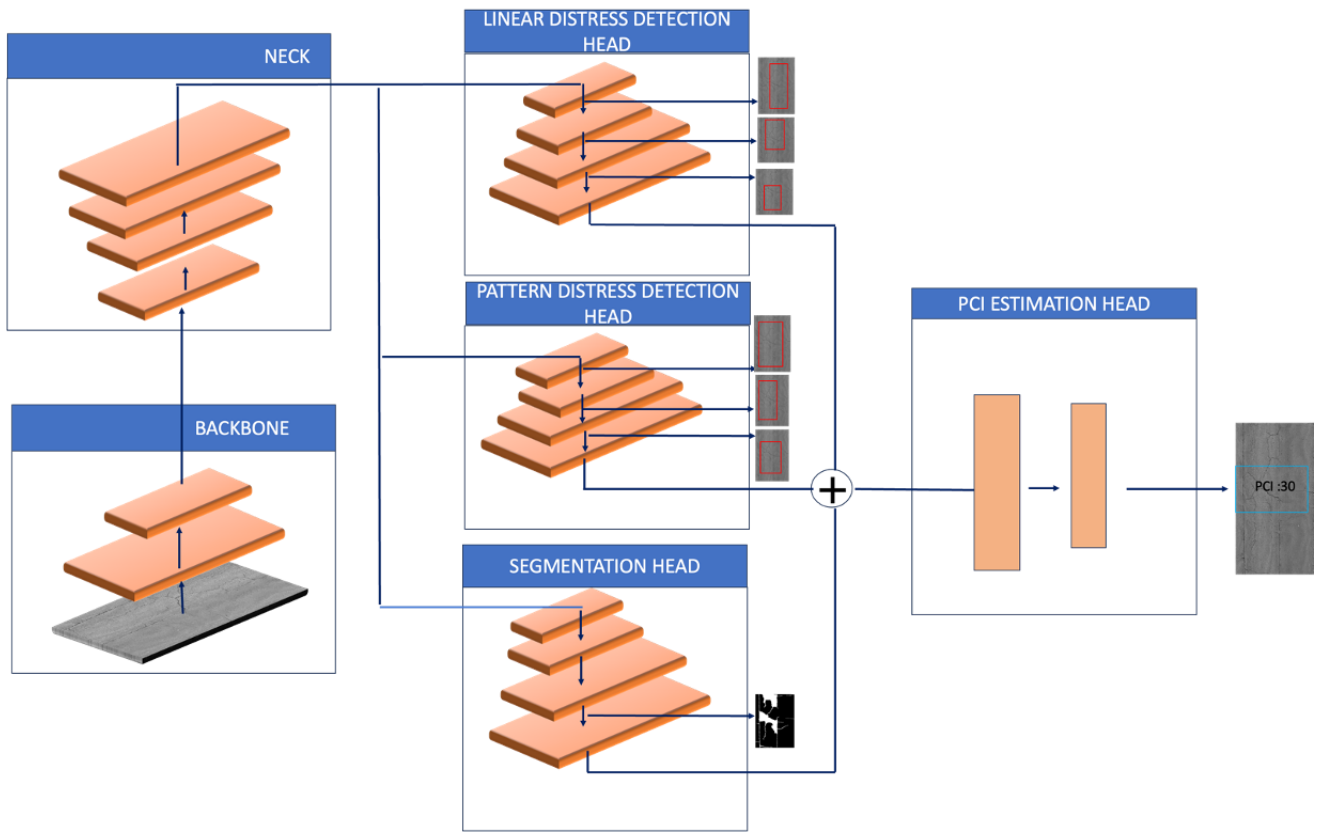

Figure 4: Model Architecture

Our detection heads were separated into linear and pattern distresses because the linear cracks (longitudinal and transverse cracks) both carry the same deduct value when estimating the PCI. On the other hand, the pattern cracks (alligator, block and patch) show advanced stages of the linear cracks and generally have higher deduct values when estimating the PCI. In the detection head, the model learns the type of distress and implicitly learns the extent --the density of the pixels within the bounding box. Therefore, from the detection head, the model implicitly learns the type and extent of the distress, two of the components used for PCI estimation. The segmentation is pixel-level classification; therefore, the model is implicitly learning the severity of the crack. Severity of linear cracks is determined by the width of the crack and if the crack is segmented, implicitly the model learns the severity of the crack. For the pattern cracks, the model can implicitly learn the severity of the distress based on the area they cover, where larger areas typically indicate more severe pattern distresses. With the features learnt from the detection head (type and extent) plus features learnt from the segmentation head(severity), these features are concatenated to form the PCI head which estimates the PCI.

**Encoder**
Image2PCI Encoder is responsible for feature extraction, and it primarily consists of the backbone for the initial feature extraction from pavement images and the neck that further processes the extracted features spatially. The backbone is comprised of CSPDarknet, which is an extension of the Darknet architecture, widely utilized in YOLO models. It employs Cross-Stage Partial connections (CSP) to enhance information flow and gradient propagation. By partitioning feature maps and introducing CSP residual blocks, CSPDarknet improves feature extraction, resulting in enhanced real-time model performance and efficiency. In our PCI estimation model, we seek efficiency and enhanced performance. The neck further processes the extracted features from the backbone, using a combination of Spatial Pyramid Pooling (SPP), BottleneckCSP blocks, convolutions, and Feature Pyramid Network (FPN). The SPP fuses multi-scale information, enhancing the model's ability to detect objects of varying sizes. The Feature Pyramid Network (FPN) incorporates feature concatenation with features from earlier stages, enabling the model to integrate information from multiple scales and produce a more robust representation of both low-level and high-level features.

**Decoder**
Image2PCI contains four decoder heads that perform four specific tasks: Linear distress detection, Pattern distress detection, Segmentation and PCI estimation.

*Linear distress detection and Pattern distress detection head*
An identical multi-scale detection head is used for both the linear distress and pattern distress heads. The bottom layer of the FPN from the neck (layer 16) is the input for the detection head. The detection layer takes in three (3) feature maps. These three layers are: 17, 20, 23 for linear distress detection head and 25, 28, 31 for Pattern distress detection head. Each feature map is assigned box configurations (box size and aspect ratios) and predictions of bounding boxes, object confidence and class probabilities are made for each feature map. The architecture demonstrates a Path Aggregation Network (PAN) approach, whereby the higher-level feature maps are fused with lower-level ones in layers 17, 20, 23 for the distress detection head and 25,28,31 for the pattern distress detection head.

*Segmentation Head*
The bottom layer of the FPN from the neck also serves as input to the segmentation head. With an input size of (W/8, H/8, 256), the segmentation head processes the input with a combination of a couple of BottleneckCSP blocks, convolutions and upsampling. This results into an output feature map of size (W, H, 2). This output feature map represents the probability of each pixel in the image being related to either pavement distress or the background.

*PCI estimation Head*
The PCI estimation head receives feature maps from two distinct sources: the final layer of the segmentation head and the layer immediately before the last layer in the detection heads (specifically, layer 23 in the linear detection head and layer 31 in the pattern detection head). These feature maps from the detection heads have an input size of (W/16, W/16, 512). To align the output channels with those from the segmentation head (W, H, 2), the detection head feature maps undergo processing through BottleneckCSP blocks, convolutions, and upsampling, resulting in an output size of (2W, 2H, 2).
Furthermore, Adaptive Average pooling is applied to the feature maps from the detection heads to bring them to (W, H, 2). The resulting feature maps from both detection heads and the segmentation head are concatenated and flattened. These combined feature maps are then passed through a Fully Connected layers to generate the PCI of the image. The concatenation of the feature maps allows the model to implicitly learn the type and extent information from the detection head and the severity information from the segmentation head. The overall architecture enables the model to effectively learn and analyze various distress patterns in pavement images, allowing for accurate prediction of the PCI, which is essential for assessing the condition of roads and making informed maintenance decisions.

**Loss function**
We use a multi-task loss such that the model can learn to estimate the PCI end-to-end. The overall loss is given by **Equation 2**:

$$Loss = \gamma_{det1}\mathcal{L}_{det1} + \gamma_{det2}\mathcal{L}_{det2} + \gamma_{seg}\mathcal{L}_{seg} + \gamma_{pci}\mathcal{L}_{pci} \qquad (2)$$

Where:

The detection head loss is a calculated from the summation of loss from the class, object and bounding box given by **Equation 3**:

$$\mathcal{L}_{det} = \beta_1\mathcal{L}_{class} + \beta_2\mathcal{L}_{obj} + \beta_3\mathcal{L}_{box} \qquad (3)$$

Segmentation loss is cross entropy loss given by **Equation 4**:
$$\mathcal{L}_{seg} = \mathcal{L}_{ce} \qquad (4)$$

and lastly PCI loss is Mean square error loss given by **Equation 5**:

$$\mathcal{L}_{pci} = \mathcal{L}_{mse} \qquad (5)$$

**RESULTS**
**Figure 5** shows the detections, segmentations and the PCI estimation of the Image2PCI model. The model effectively segments the pavement distresses and ably distinguishes a linear pavement distress from the pattern distress. Additionally, the model's PCI estimation shows a high degree of accuracy.

| Actual:50 | Actual:99 | Actual:65 |
| --- | --- | --- |
| Predicted: 60 | Predicted: 97 | Predicted: 69 |

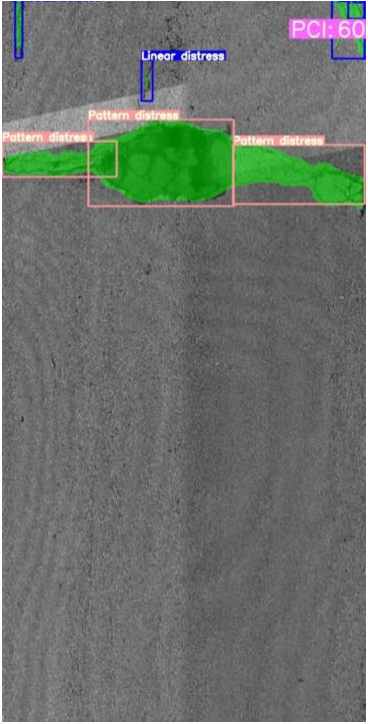 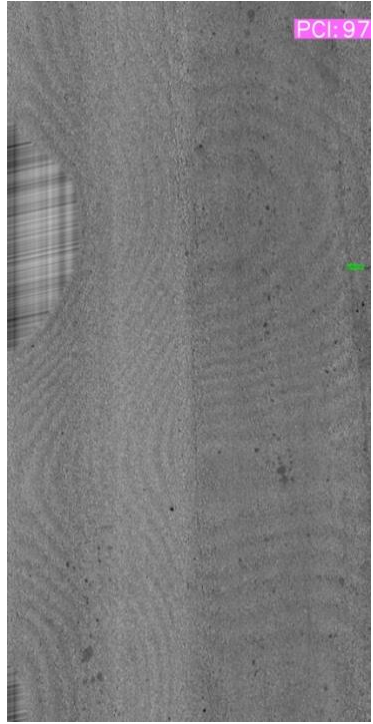 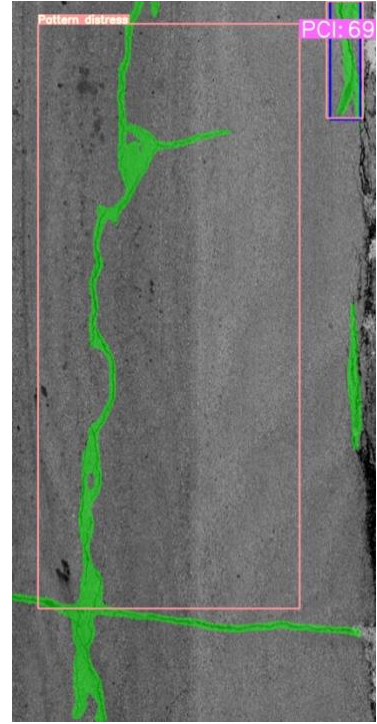

| Actual:78 | Actual:91 | Actual:99 |
| --- | --- | --- |
| Predicted:79 | Predicted: 95 | Predicted:98 |

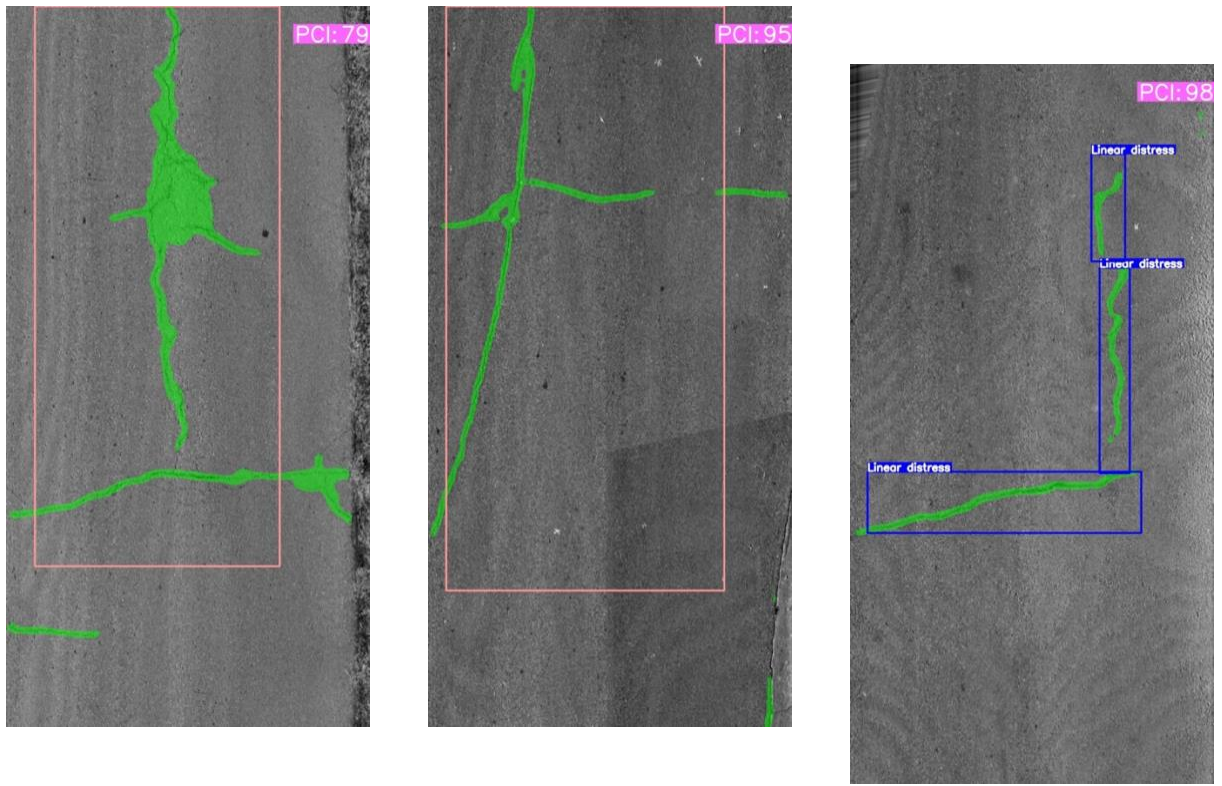

**Figure 5: Image results of Image2PCI model**

**Wrong PCI estimations**
The model missed some detections and gave wrong predictions of PCI as shown in **Figure 6**. This could be attributed to the skewed representations of the PCI values in the training data as shown on **Figure 7.**

| Actual: 27    | Actual:58     | Actual:51     |
|---------------|---------------|---------------|
| Estimated: 65 | Estimated: 80 | Estimated: 82 |

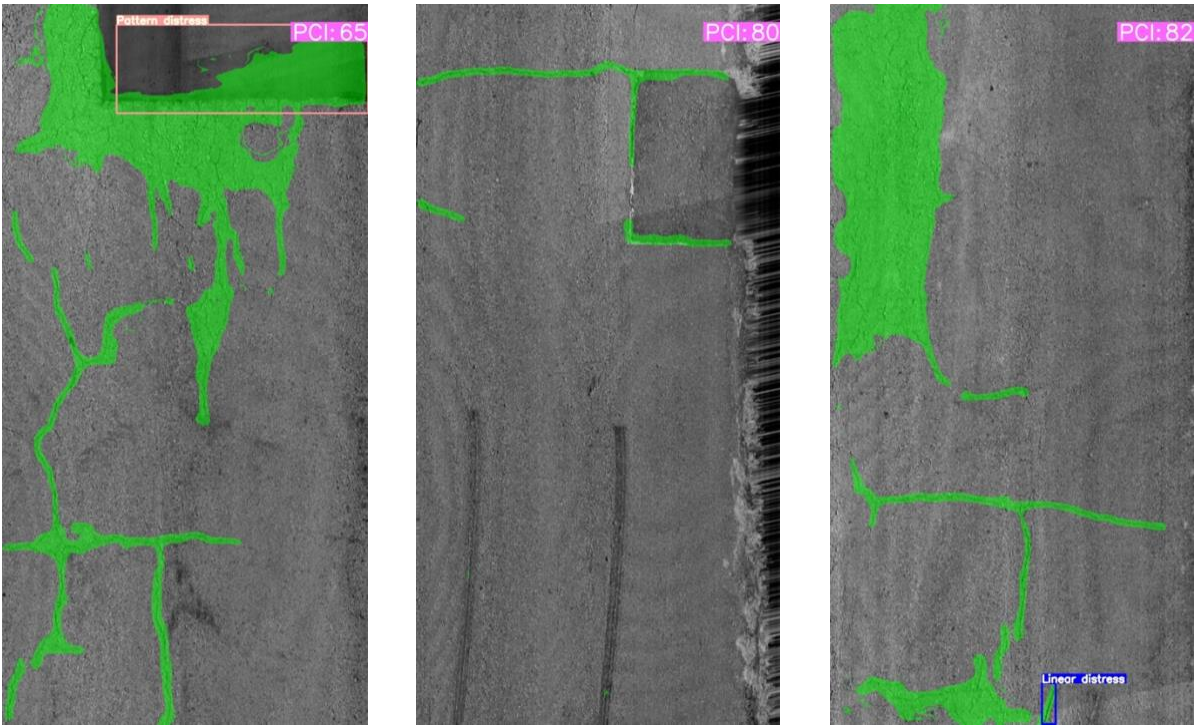

**Figure 6: Wrong PCI predictions**

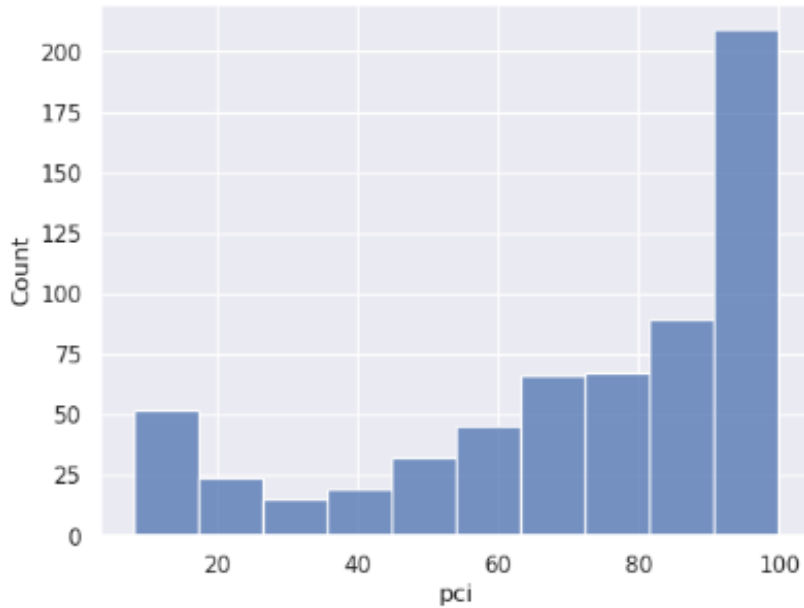

**Figure 7: Histogram of PCI values in training data**

To gain a better understanding of the model's performance, the actual PCI and the predicted PCI are plotted as seen in **Figure 8.** The green line represents the actual values, and the yellow line represents the estimated PCI. From this figure we can see the predicted results have a relatively small error.

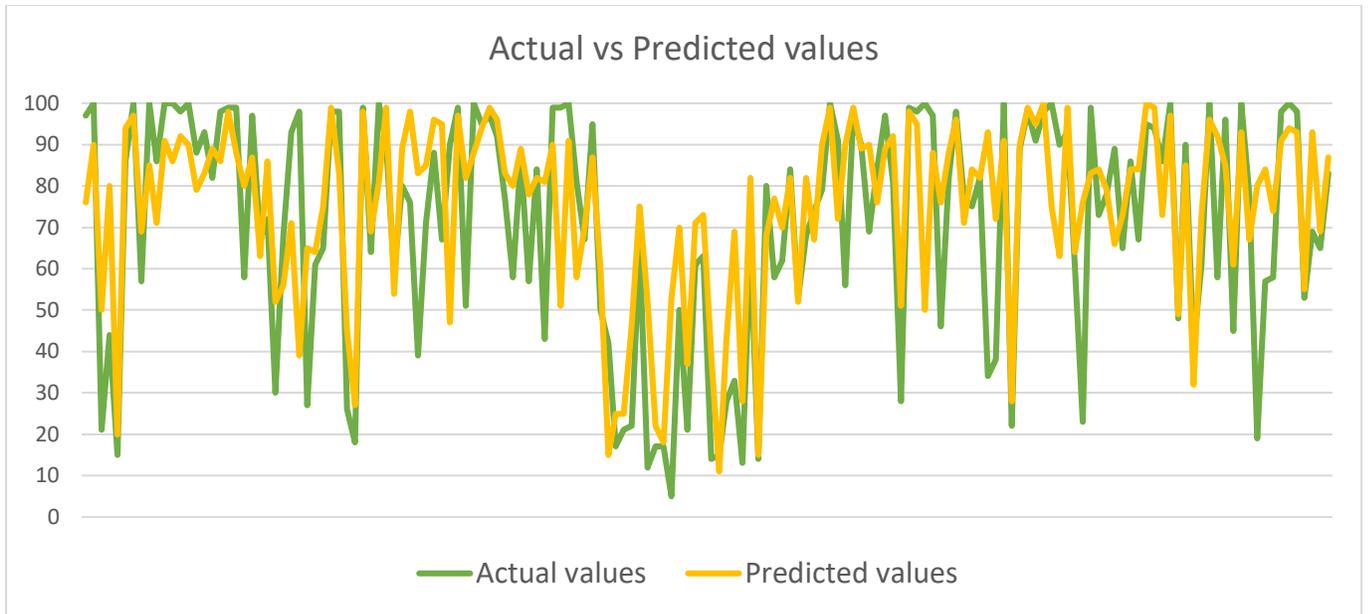

**Figure 8: Actual PCI vs Predicted PCI**

Furthermore, to understand the correlation between the predicted and actual PC values, $R^2$ and MAPE are calculated as shown in **Figure 9. Figure 9** shows a line graph of the Predicted PCI vs the Actual PCI values. The vertical axis represents the Predicted PCI value and the horizontal axis represents the actual value. The correlation of the predicted PCI and ground truth has $R^2$ of 0.75 and MAPE of 10.4%. This means the predictions are off by 10.4%

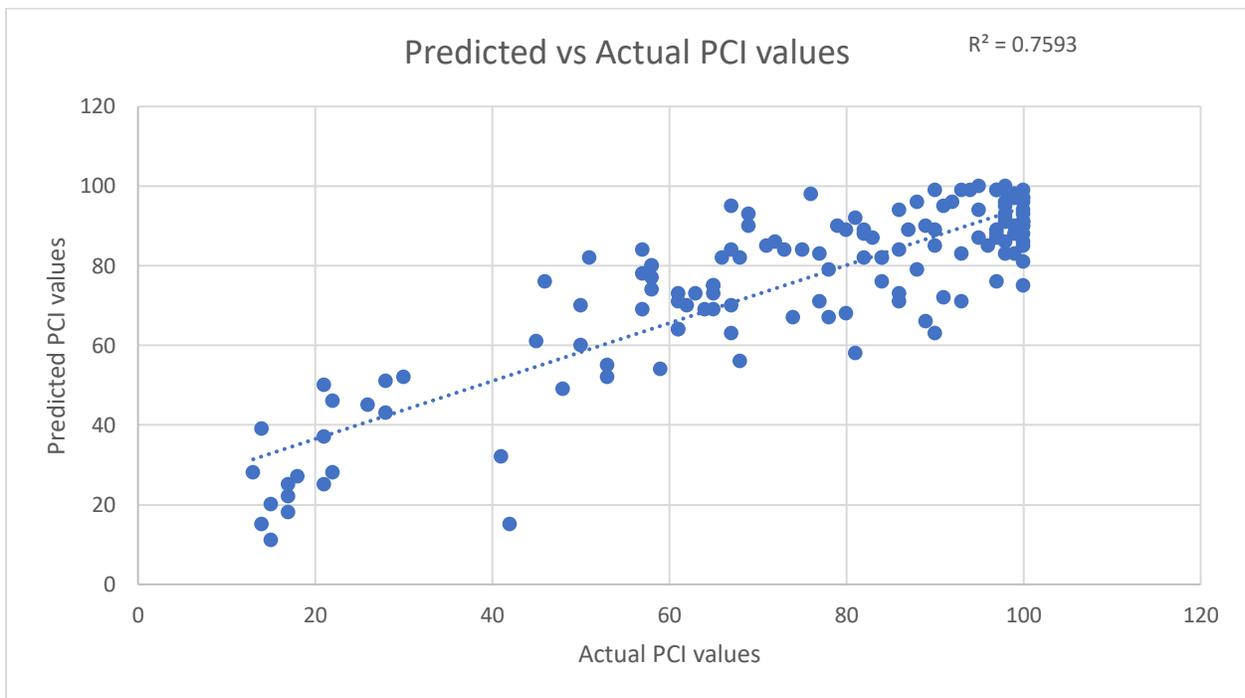

**Figure 9: Predicted vs Actual PCI**

Image2PCI is the very first of its kind to estimate PCI directly from images. The model performs very well on our benchmarked and open pavement distress dataset giving an $R^2$ of 0.75 and an MAPE of 10.5%

**CONCLUSIONS**

The Pavement Condition Index (PCI) is a widely used metric for evaluating pavement performance based on the type, extent and severity of distresses detected on a pavement surface. The field of artificial intelligence, fueled by advancements in deep learning, has increased the demand for accurate and cost-effective network-level pavement monitoring. Diverse studies have investigated deep-learning distress analysis algorithms to facilitate automated pavement evaluation. Nevertheless, the most common method employs separate models for various tasks, including detection, segmentation, and severity estimation. Some studies, for instance, combine Yolov4 and Deep Labv3 models for detection and segmentation, whereas others use Mask RCNN and image processing for crack detection and severity evaluation. Despite the fact that these two-step procedures yield precise Pavement Condition Index (PCI) estimates, they have significant drawbacks. They are computationally expensive, rendering them impractical for deployment on edge devices, and they lose opportunities to leverage shared features between related tasks, limiting generalization and overall accuracy.

The authors proposed a novel multitasking model capable of implicitly learning pavement distress type, extent and severity estimation simultaneously in a single phase to overcome the limitations of existing two-step approaches. By integrating these closely related tasks into a unified framework, the proposed model learns from shared characteristics, allowing for more efficient and precise pavement condition assessment. This method not only reduces computational costs, making it suitable for deployment on edge devices, but it also maximizes the utilization of information from each task, resulting in enhanced generalization and accuracy across multiple related tasks. The goal of this study is to develop a single multitask model that can estimate PCI directly from an image and introduce a new benchmark of annotated datasets with polygon-based annotations that combine both the type and severity of cracks for training and benchmarking unified Pavement Condition Index (PCI) estimation models.

From the study, the model was able to efficiently and effectively estimate PCI directly from the images with a $R^2$ of 0.75 and an MAPE of 10.4%. To our best knowledge, this is the first work that can estimate PCI directly from an image at real time speeds while maintaining excellent accuracy on all related tasks for crack detection and segmentation. The study's limitation is attributed to the time-consuming nature of the polygon-based annotations consisting of both type and severity, resulting in a small amount of data for training and testing.

The authors believe that this study and dataset created will contribute greatly to conducting automated comprehensive, network level PCI surveys. Further studies could be done to improve the accuracy of the model by increasing the number of images annotated and enhancing the model architecture.

**AUTHOR CONTRIBUTIONS**
The authors confirm contribution to the paper as follows: study conception and design: N.J. Owor, Y. Adu-Gyamfi; data annotation: H. Du, A. Daud; analysis and interpretation of results: N.J. Owor, Y. Adu-Gyamfi; draft manuscript preparation: N.J. Owor, A. Daud, Y. Adu-Gyamfi. All authors reviewed the results and approved the final version of the manuscript.